\pdfoutput=1

\documentclass[11pt]{article}

\usepackage[]{acl}

\usepackage{times}
\usepackage{latexsym}
\usepackage[T1]{fontenc}

\usepackage[utf8]{inputenc}

\usepackage{microtype}
\usepackage{graphicx}
\usepackage{multirow}
\usepackage{stfloats}
\usepackage{pdfpages}
\usepackage{algorithm}
\usepackage{algorithmic}
\usepackage{times}

\usepackage{soul}
\usepackage{url}
\usepackage[utf8]{inputenc}
\usepackage{makecell}
\usepackage{fdsymbol}
\usepackage{verbatim}
\usepackage{microtype}
\usepackage{amsmath}
\usepackage{tabularx}
\usepackage{graphicx}
\usepackage{multirow}
\usepackage{textcomp}
\usepackage{subcaption}
\usepackage{times}
\usepackage{latexsym}
\usepackage{microtype}
\usepackage{natbib}
\usepackage{booktabs}
\usepackage{appendix}
\usepackage{tikz}
\usepackage{tikz-dependency}
\usepackage{amsmath}
\usepackage{amsfonts}
\usepackage{graphicx}
\usepackage{amsthm}
\usepackage{mathtools}
\usepackage{multirow}
\usepackage{url}
\usepackage{color}
\usepackage{dsfont}
\definecolor{color1}{rgb}{0.22,0.45,0.70}  
\definecolor{color2}{rgb}{0.45,0.45,0.45}  

\title{MoCoSA: Momentum Contrast for Knowledge Graph Completion with Structure-Augmented Pre-trained Language Models}

\author{Jiabang He\textsuperscript{\rm 2}\Thanks{This is work done during an internship at Kwai}, Liu	Jia\textsuperscript{\rm 1}, Lei	Wang\textsuperscript{\rm 3}, Xiyao Li\textsuperscript{\rm 1}, Xing Xu\textsuperscript{\rm 2} \\
       \textsuperscript{\rm 1}Kuaishou Technology, China\\ \textsuperscript{\rm 2}University of Electronic Science and Technology of China, China \\ \textsuperscript{\rm 3}Singapore Management University, Singapore \\
       JiaBangH@outlook.com, liujia08@kuaishou.com,
       lei.wang.2019@phdcs.smu.edu.sg, \\
       lixiyao@kuaishou.com,
       xing.xu@uestc.edu.cn
       }

\begin{document}
\maketitle

\begin{abstract}
Knowledge Graph Completion (KGC) aims to conduct reasoning on the facts within knowledge graphs and automatically infer missing links. Existing methods can mainly be categorized into structure-based or description-based.
On the one hand, structure-based methods effectively represent relational facts in knowledge graphs using entity embeddings. 
However, they struggle with semantically rich real-world entities due to limited structural information and fail to generalize to unseen entities. On the other hand, description-based methods leverage pre-trained language models (PLMs) to understand textual information. They exhibit strong robustness towards unseen entities. However, they have difficulty with larger negative sampling and often lag behind structure-based methods. 
To address these issues, in this paper, we propose 
\textbf{Mo}mentum \textbf{Co}ntrast for knowledge graph completion with \textbf{S}tructure-\textbf{A}ugmented pre-trained language models (\textbf{MoCoSA}), which allows the PLM to perceive the structural information by the adaptable structure encoder. To improve learning efficiency, we proposed momentum hard negative and intra-relation negative sampling.
Experimental results demonstrate that our approach achieves state-of-the-art performance in terms of mean reciprocal rank (MRR), with improvements of 2.5\% on WN18RR and 21\% on OpenBG500.

\end{abstract}

\begin{figure}[!htb]
  \centering
  \includegraphics[width=1.0\linewidth]{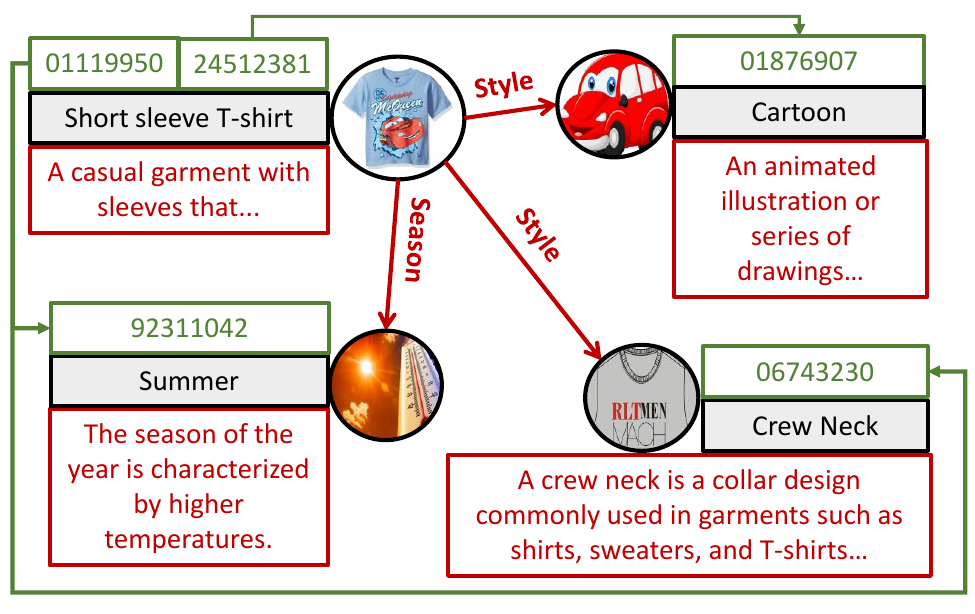}
  \caption{An illustration of knowledge graphs, each entity includes a name, description, and unique entity ID. The entity ID is highlighted in green, while the description is denoted in red and the entity name in black. 
  }
  \label{fig:abstract}
\vspace{-5pt}
\end{figure}

\section{Introduction}

Knowledge Graphs (KGs) exhibit intricate graph structures and encompass a wealth of factual textual information. Prominent public knowledge graphs encompass Wikidata~\cite{Vrandei2014WikidataAF}, YAGO~\cite{Suchanek2007YagoAC},NELL~\cite{Carlson2010TowardAA} and WordNet~\cite{Fellbaum2000WordNetA}. Nevertheless, human-crafted knowledge graphs encounter issues of incompleteness. This limitation inevitably restricts the practical applications of human-crafted knowledge graphs. Knowledge graph completion (KGC) aims to automatically complete triples with missing entities in the knowledge graph. In this paper, we focus on the link prediction task in KGC, which aims to predict the missing head (or tail) entity when provided with the relation and tail (or head) entity in a triple.

\begin{figure*}[!htb]
  \centering
  \includegraphics[width=1.0\linewidth]{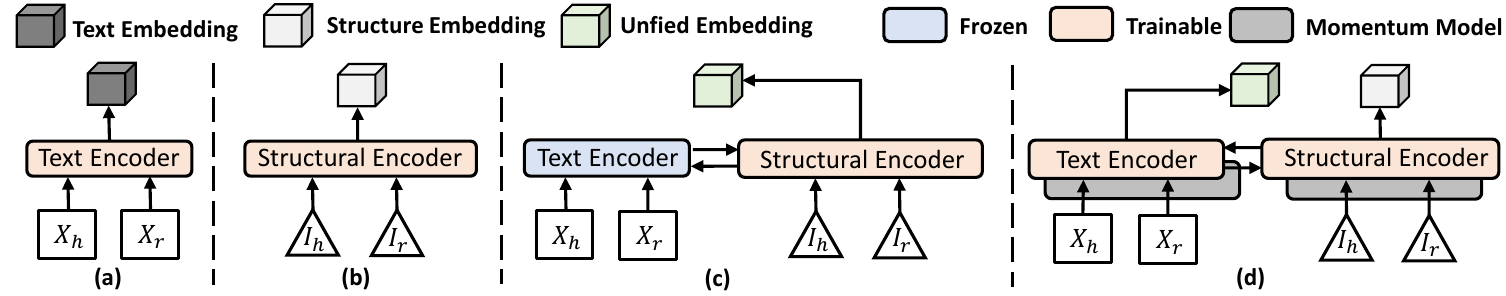}
  \caption{(a) The description-based methods use PLM to convert head ($X_{h}$) and relation ($X_{r}$) to text embeddings. (b) The structure-based methods represent the head index ($I_{h}$) and the relation index ($I_{r}$) as embeddings to learn the structural information.
  (c) CSPromp-KG~\cite{Chen2023DippingPS} integrates both types of information using a soft prompt.
  (d) Our MoCoSA employs the structure-augmented PLM to learn unified embedding and structure embedding.
  }
  \label{fig:method_1}
\vspace{-10pt}
\end{figure*}

Existing KGC models can be categorized into structure-based or description-based. The structure-based KGC methods~\cite{bordes2013translating, trouillon2016complex} employ trainable structure embedding for entities and relations. These models are trained to preserve connections between head entities and relations to infer tail entities through various transitional relationships. Although these methods are effective in modeling KG structural information, 
they are unable to infer entities that are not present in the training set. As shown in Figure~\ref{fig:abstract}, textual entity descriptions encompass a wealth of information. The description-based KGC methods~\cite{Choi2021MEMKGCME, yao2019kg, Wang2022LanguageMA} proposed fine-tuning the pre-trained language models (PLMs) to represent the entities and relations based on textual context. PLMs provide sufficient robustness for KGC based on textual information, but they often lag behind structure-based methods on popular benchmarks. To better exploit the description-based method for KGC, drawing inspiration from strategies in visual representation learning~\cite{Khosla2020SupervisedCL} and dense passage retrieval~\cite{Karpukhin2020DensePR}, SimKGC~\cite{Wang2022SimKGCSC} propose a contrastive learning framework
 which incorporates three distinct types of negatives: in-batch negatives, pre-batch negatives, and self-negatives. While SimKGC stands as the first description-based KGC method to outperform the structure-based counterparts, it still grapples with certain challenges. We categorize the two reasons as follows: on the one hand, in structure-based methods like RotatE~\cite{sun2018rotate}, a triple can be modeled as rotation in complex space or relational translation, while SimKGC
does not enable such easy-to-understand interpretations. On the other hand, negative sampling strategies exhibit certain drawbacks: (1) The in-batch negatives are coupled with batch size. (2) The pre-batch negatives, only 1 or 2 pre-batches are used which does not provide sufficient negatives. (3) The Self-negatives employ the head entity as a challenging negative yet in instances such as (`Dress', `Style Name', `Dress') found within OpenBG500, the head entity could actually be a positive. Thus, it is imperative to continue exploring effective methodologies for mitigating false negatives.

It is natural to integrate both contextualized and structured knowledge into a unified model for joint learning and to propose more effective negative sampling strategies. CSPromp-KG~\cite{Chen2023DippingPS} employs parameter-efficient
prompt learning for KGC, which is a description-based KGC model that effectively fuses the KG structural knowledge and avoids over-fitting towards textual information. Additionally, it proposes local adversarial regularization (LAR) entities for negatives but is limited by a small number for each training instance. The challenge of introducing more potent negative samples to enhance learning efficiency remains an open issue.
Therefore, we focus on the research question in this paper: \textit{Can we integrate both the contextualized and structured knowledge for knowledge graph completion, while proposing more effective negatives?}


To this end, we propose Momentum Contrast for Knowledge Graph Completion with Structure-Augmented Pre-trained Language Models (MoCoSA), a novel method for KGC with a contrastive learning framework. As shown in Figure \ref{fig:method_1}, compared with other model architecture, We independently encode the textual and structural representations and then fuse them through the text encoder for learning the unified embeddings.
We integrate the structural information in the text encoder, which serves two purposes:
(1) Balancing the learning of textual and structural information. (2) Learning a shared dimensional space to embed textual and structural information, which matches embeddings to discover more informative samples through the negative sampling strategy. Additionally, we introduce two types of mining hard negatives to improve the contrastive learning efficiency: momentum hard negatives and instra-relation negatives. Vectors from previous batches are cached in a momentum queue \cite{he2020momentum} and mix \cite{Kalantidis2020HardNM} some of the
hardest negative features of the contrastive loss in the momentum queue as momentum hard negatives. Additionally, for a given triple, the other entity in the candidate tail entities linked to this relation can be served as hard negatives, which we refer to as instra-relation negatives.

The utilization of existing true triples in KGs is of paramount importance. To address the tail entity recall limitations while predicting, we propose a simple and effective relation-based re-ranking strategy. It involves reducing the score of the non-tail neighbors based on the relation link by using the information from the existing true triples.

We evaluate our proposed model MoCoSA by conducting experiments on three popular benchmarks: WN18RR \cite{Dettmers2017Convolutional2K}, FB15k-237 \cite{Ma2015KnowledgeGI}, and OpenBG500 \cite{Deng2022ConstructionAA}. MoCoSA consistently outperforms existing baseline models on WN18RR (MRR 67.1\% $\rightarrow$ 69.6\%) and OpenBG500 (MRR 42.7\% $\rightarrow$ 63.4\%). We carry out an ablation study to demonstrate the effectiveness of individual components. Additionally, we replace different structure-based methods to evaluate the flexibility of adaptable structural encoders.

\section{Related Work}

\subsubsection{Structure-based methods.}

Knowledge Graph Embedding (KGE) aims to map each entity and its relation to low-dimensional vector spaces as embeddings. For each triplet ($h$, $r$, $t$), they model the triplet's plausibility using a distance-based scoring function $f$($h$, $r$, $t$). This function measures the distance between the entity embeddings of $h$ and $t$, specific to the relation $r$. The most famous method is TransE~\cite{bordes2013translating}, which embeds entities and relations in a shared vector space of dimension $d$. The loss function is defined as $\left\Vert \textbf{h}+\textbf{r}-\textbf{t} \right\Vert$, aiming to bring $h$ close to $t$ after translation by $r$. 
It can be extended using different geometric transformations, such as TransH~\cite{Wang2014KnowledgeGE}, which projects entity embeddings of $h$ and $t$ onto relation-specific hyperplanes, or RotateE~\cite{sun2018rotate}, which defines relation as a rotation from entity $h$ to entity $t$ in a complex vector space. Therefore, their embeddings are expected to satisfy $\textbf{h}\odot\textbf{r}\approx\textbf{t}$, where $\odot$ denotes element-wise multiplication. 
DistMult~\cite{yang2014embedding} proposes to restrict the relation matrix to a diagonal matrix, significantly reducing the number of parameters in the bilinear model. However, these methods lack the integration of entity descriptions, thus displaying a limitation in their capacity to generalize to entities that have not been seen during training.

\subsubsection{Description-based methods.}

The integration of Large Language Models (LLMs) has recently facilitated KGC methods in encoding or generating facts from textual information. KG-Bert~\cite{yao2019kg} represents triples as text sequences and encodes them using a text encoder like BERT~\cite{devlin-etal-2019-bert}. SimKGC~\cite{Wang2022SimKGCSC} employs text encoders to generate textual representations, which are subsequently enhanced using contrastive learning techniques. This process needs to compute the similarity between the positive and negative samples. More specifically, it aims to maximize the similarity between the positive sample, while minimizing the similarity between the negative sample. In contrast to prior models focusing solely on encoding, recent studies have employed LLMs as sequence-to-sequence generators in KGC. ~\cite{Xie2022FromDT, Saxena2022SequencetoSequenceKG, Xie2022FromDT}. When handling closed-source LLMs such as GPT-3.5~\cite{Chen2020BigSM} and GPT-4 ~\cite{OpenAI2023GPT4TR}, AutoKG employs prompt engineering to design customized prompts~\cite{Zhu2023LLMsFK}, which guide LLMs to generate the tail entities in KGC. Although these methods leverage the robustness of PLMs in handling textual information of entities, they still fall behind structure-based methods and lack effective negative sampling strategies.

\section{Our MoCoSA Model}

In this section, we first introduce Knowledge Graph Completion (KGC), Then we provide a detailed description of the structure-augmented module for PLM and the adaptable structural encoder. In addition, we propose negative sampling based on momentum negatives and intra-relation negatives. Finally, we present the relation-based re-ranking strategy, as well as the training and inference.

\subsection{Problem Definition: Knowledge Graph Completion}

Given a specific triple $u=f(h,r,t)$ from the knowledge graph $\mathcal{G}$, where $h, t\in \mathcal{E}$ and $r\in \mathcal{R}$. $\mathcal{E}$ is the set of entites and $\mathcal{R}$ is the set of relations. The evaluation protocols utilize entity ranking as part of the assessment process. Tail entity prediction ($h$, $r$, $?$) entails ranking all entities associated with given $h$ and $r$, similarly for head entity prediction ($?$, $r$, $t$). Following the approach presented in SimKGC \cite{Wang2022SimKGCSC}, we use the inverse triples ($t, r^{-1}, h$), where $r^{-1}$ denotes the inverse relation of $r$, simplify the task by focusing exclusively on tail entity prediction.

\subsection{Model Architecture}

\begin{figure*}[!htb]
  \centering
  \includegraphics[width=1.0\linewidth]{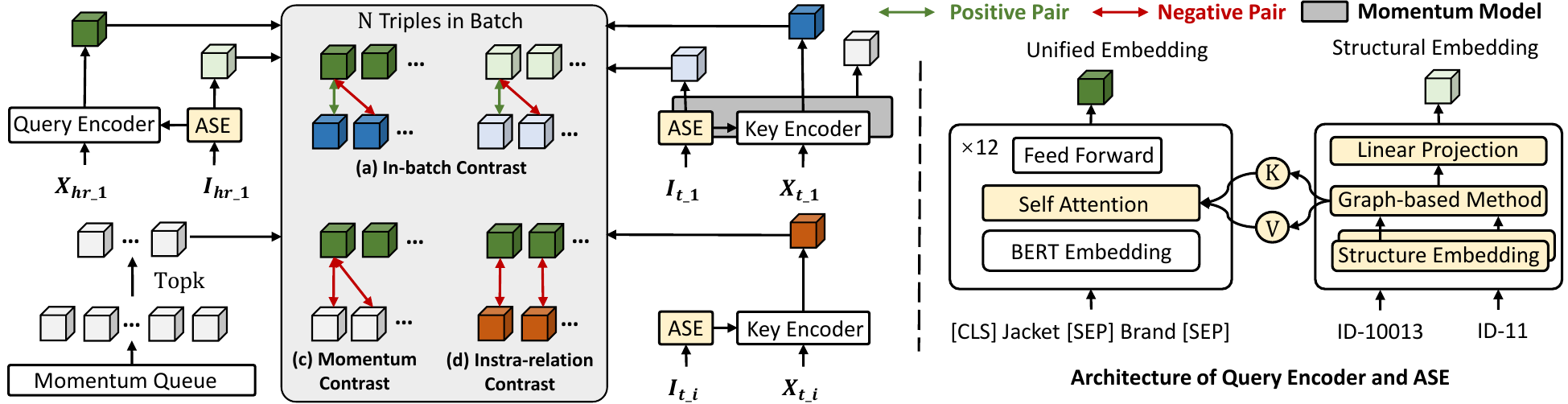}
  \caption{An overview of the framework of MoCoSA, with the model architecture of structure-augmented PLM shown on the right. ASE is the adaptable structural encoder. The structural features interact with the textual query in the self-attention mechanism of PLM. To improve the performance of MoCoSA, we use three contrastive losses named in-batch 
 contrast, momentum contrast, and instra-relation contrast are shown on the left.}
  \label{fig:method_2}
\vspace{-5pt}
\end{figure*}

As illustrated in Figure~\ref{fig:method_2}, the proposed MoCoSA contains a text encoder and an additional structural encoder. For the text encoder, we adopt two 12-layer textual transformers (i.e., BERT~\cite{devlin-etal-2019-bert}) as our pre-trained models. 
The text encoders are initialized by weights of the BERT$_{\text{BASE}}$. For the structure encoders, we compose an embedding layer and a linear projection, which is initialized with random weights. The hyper-parameter details of these encoders are described in the Experiment.

\subsection{Structure-Augumented Module for PLM}

Formally, for a given triplet ($h$, $r$, $t$), the head entity with the relation and the tail entity are represented by their respective natural language descriptions and unique indices. MoCoSA leverages the description $d_{h}, d_{r}, d_{t}$ and the index $I_{h}, I_{r}, I_{d}$ to serve as additional inputs. Then constructed head entity with relation is represented in the following input sequence format: $x_{hr} = ([CLS], h, d_{h}, [SEP], r, d_{r}, [SEP])$. BERT$_{\text{hr}}$ is employed to extract the hidden state $h_{hr}$ from the last layer. Instead of utilizing the hidden state solely from the [$CLS$] token, we adopt a mean pooling technique followed by $\ell_{2}$-norm to derive textual relation-aware embeddings, which has been shown to result in better sentence embeddings~\cite{gao2021simcse,Reimers2019SentenceBERTSE}. 
Subsequently, we prepend structural information to BERT$_{\text{hr}}$ to acquire, where we use the structural embeddings $\mathbf{E_{h}}\in \mathbb{R}^w$ for the head index $I_{h}$ and $\mathbf{E_{r}}\in \mathbb{R}^w$ for the relation index $I_{r}$ to interact in BERT transformer layers. $w$ is the embedding dim. Formally,
\begin{equation}
    \mathbf{E_{hr}}=\mathbf{W_{out}} \cdot ASE(\mathbf{E_{h}}, \mathbf{E_{r}}),
\end{equation}
\begin{equation}
    \mathbf{K}'_{LM}=[\mathbf{E_{hr,k}};\mathbf{K_{LM}}],  \mathbf{V}'_{LM}=[\mathbf{E_{hr,v}};\mathbf{V_{LM}}].
\end{equation}
where $ASE(\cdot)$ denotes the adaptable structural encoder. $\mathbf{W_{out}}\in \mathbb{R}^{w \times (2 \cdot l \cdot d)}$. $\mathbf{E_{hr,k}}$, $\mathbf{E_{hr,v}}$ are computed through linear transformations on $\mathbf{E_{hr}}$. 
The query ($\mathbf{Q_{LM}}$), key ($\mathbf{K_{LM}}$), and value ($\mathbf{V_{LM}}$) are computed linear transformations on the hiddens states of $h_{hr}$. The structure-augmented self-attention layer is finally calculated as follows:
\begin{equation}
\begin{aligned}
    \mathbf{A_{self}}=\operatorname{softmax}(\mathbf{Q_{LM}}{\mathbf{K}'^{\top}_{LM}})\mathbf{V}'_{LM}.
\end{aligned}
\end{equation}
Finally, we obtained structure-augmented text features $\mathbf{h_{hr}}$ for the head entity $h$ and relation $r$ after computing the hidden state through $\mathbf{A_{self}}$. 
Similarly, we can get $\mathbf{h_{t}}$ abd the structural embedding $\mathbf{E_{t}}$ for the tail entity $t$. There exists a comprehensive interaction, leading to the fusion of structural knowledge extracted from $\mathcal{G}$ and textual knowledge derived from PLM. Given that the embeddings $\mathbf{h_{hr}}$ and $\mathbf{h_{t}}$, the cosine similarity can be simply computed as the dot product of these two embeddings:
\begin{equation}
    cos(\mathbf{h_{hr}}, \mathbf{h_{t}}) =\frac{\mathbf{h_{hr}} \cdot \mathbf{h_{t}}}{||\mathbf{h_{hr}}|| ||\mathbf{h_{t}}||} = \mathbf{h_{hr}} \cdot \mathbf{h_{t}}.
\label{cossim}
\end{equation}
For tail entity prediction, we calculate the cosine similarity between $\mathbf{h_{hr}}$ and all entities in $\mathcal{G}$, selecting the one with the highest score as the prediction.
\begin{equation}
\arg\max\limits_{t_{i}\in \mathcal{E}}{cos(\mathbf{h_{hr}}, \mathbf{h_{t_{i}}})}.
\end{equation}
\subsection{Adaptable Structural Encoder}
It is possible to improve knowledge acquisition in structure-based models for more precise structural information transmission during contrastive learning. Translation-based graph embedding methods commonly utilize structural encoder $SE(\cdot)$ for modeling spatial distance function $d$. In particular, TransE~\cite{bordes2013translating} assign a score function $d(\mathbf{h},\mathbf{r},\mathbf{t})$ inversely proportional to the spatial distance: 
\begin{equation}
SE(\mathbf{h}, \mathbf{r})=\mathbf{h} + \mathbf{r},
\end{equation}
\begin{equation}
d(h, r, t)=-\lVert SE(\mathbf{E_{h}}, \mathbf{E_{r}})-\mathbf{E_{t}} \rVert. 
\end{equation}
Structured knowledge is acquired by maximizing the score margin between positive triples and negative triples. Here, We adopt a method to constrain the distance in triples, denoted as the adaptable structural encoder $ASE(\cdot)$ for different structure-based methods. We use the cosine similarity measures instead of distance functions and optimize the probability-based structural loss to reconstruct the structural information transferred to the textual information. The score function as follows:
\begin{equation}
\begin{aligned}
d(h, r, t) = cos(ASE(\mathbf{E_{h}}, \mathbf{E_{r}}), \mathbf{E_{t}})).
\end{aligned}
\end{equation}
\subsection{Momentum Hard Negative Sampling}
In-batch ($\textbf{IB}$) negative sampling \cite{Wang2022SimKGCSC} serves negative examples that entities within the same batch. These negative sampling technologies enable the efficient use of entity embeddings for bi-encoder models but are dependent on the training batch size. We follow this negative sampling strategy and propose increasing the number of negative samples during the contrastive learning process named momentum hard (\textbf{MH}) negative sampling. Inspired by MoCo~\cite{he2020momentum}, we use the momentum queue $Q$ to store the recent $M$ tail features $p$ that are generated by the momentum key encoder with ASE. To enhance the quality of features through real-time computations with minimal computational overhead, we mixup~\cite{Kalantidis2020HardNM} the challenging negative features to generate $\tilde{h}_{t}$. 
\begin{equation}
\begin{aligned}
    \mathbf{\tilde{h}_{t}}=\frac{\mathbf{\tilde{p}}}{\Vert \mathbf{\tilde{p}} \Vert_2}, \ where \
    \mathbf{\tilde{p}} = \lambda_{k} \mathbf{p_{i}} + (1 - \lambda_{k})\mathbf{p_{j}}.
    \end{aligned}
\end{equation}
where $\lambda_{k} \in (0, 1)$ is a randomly selected mixing coefficient and $\Vert \cdot \Vert_2$ is the $\ell_{2}$-norm. $\mathbf{p_{i}}, \mathbf{p_{j}}$ are randomly selected from $\tilde{Q}^{k}$ which is the closest set from the sorted momentum queue $\tilde{Q}=\left\{\mathbf{p_{1}},\mathbf{p_{2}},..,\mathbf{p_{M}}\right\}$ such that: $\mathbf{p_{i}}> \mathbf{p_{j}}, \forall i < j$. $\tilde{Q}$ is descended by logits $l(\mathbf{p})=\mathbf{h_{hr}}^T \cdot \mathbf{p}/\tau$, where $\tau$ is a learnable temperature parameter.

\subsection{Intra-relation Negative Sampling}

Most existing methods randomly corrupt either $h$ or $t$ and then filter out the false negatives, which appear in the training graph $\mathcal{G}$. However, the negatives for different triples are independent.~\cite{sun2018rotate, Wang2020StructureAugmentedTR}. Description-based methods often assign high scores to candidate tail entities linked to the different relation $r$, which might be due to the high textual similarity of entities. To mitigate this concern, We propose intra-relation (\textbf{IR}) negative sampling for tail entities to construct corresponding wrong triples $tp'$:
\begin{equation}
	IRNS(tp') = \left\{(h,r,t') | t' \in e(r) - e(h,r) \in \mathcal{G}\right\}.
\end{equation}
Where $'$ means that are derived from negative examples, $e(r)$ represents the set of tail entities that are linked to the relation $r$ in $\mathcal{G}$, and $e(h,r)$ represents the set of tail entities that are linked to by the combination of entity $h$ and relation $r$ in $\mathcal{G}$.

\subsection{Relation-based Re-ranking}

The utilization of existing true triples in $\mathcal{G}$ is also crucial. TransE~\cite{bordes2013translating} proposes to ignore the scores of all known true triples during training, validation, and test set. To compensate for MoCoSA's tail entity recall in learning the same relation, we present a simple and effective relation-based re-ranking strategy: based on the existing true triples, if the tail entity $t_{i}$ does not exist in tail neighbors $\mathcal{N}_{r}$ that linked to by the relation $r$, the score of the tail entity $t_{i}$ will be decreased by $\alpha \ge 0$.
\begin{equation}
\underset{t_{i} \in \mathcal{E}}{\mathrm{argmax}}\cos (\mathbf{h_{hr}}, \mathbf{h_{t_{i}}}) - \alpha\mathds{1}{1}(t_{i} \notin \mathcal{N}_{r}).
\end{equation}

We use the relation-based re-ranking by calculating the mean average intersection over union (MIOU) ratio of the tail entities, which are linked to relations present in the training $(tr)$ and test $(te)$ sets. It can be formulated as follows:
\begin{equation}
MIOU=\frac{1}{N}\sum_{i=1}^N\frac{e(r_{i_{tr}}) \cap e(r_{i_{te}})}{e(r_{i_{tr}}) \cup e(r_{i_{te}})}.
\end{equation}
where $N$ means the number of triples in the test set. $e(r_{i})$ represents the set of tail entities that are linked to the relation $r_{i}$ in the specific dataset. We conduct an in-depth analysis of the application scenarios of this strategy in the experiment.

\subsection{Training and Inference}
We formulate our contrastive learning loss function based on the structure-augmented text features as follows.
\begin{equation} \label{eq:infonce}
    \mathcal{L}_{cl} = -\log \frac{e^{(cos(\mathbf{h_{hr}}, \mathbf{h_{t}})-\gamma)/\tau}} {e^{(cos(\mathbf{h_{hr}}, \mathbf{h_{t}})-\gamma)/\tau} + \mathcal{S}_{neg}},
\end{equation}
\begin{equation} \label{eq:infonce}
    \mathcal{S}_{neg} = \sum_{m=1}^Me^ {cos(\mathbf{h_{hr}},\mathbf{\tilde{h}^{m}_{t}})/ \tau}+\sum_{(h,r,t') \in \mathcal{D}}e^{cos(\mathbf{h_{hr}},\mathbf{h_{t'}})/\tau}.
\end{equation}
Where $\mathcal{D}=(IB \cup IR)$. The additive margin $\gamma > 0$ encourages the model to increase the score of the correct triples. $\tau$ is a learnable temperature parameter. To enhance the structural features, we use the in-batch (\textbf{IB}) structural features generated by the adaptable structural encoder. The structural contrastive loss $\mathcal{L}_{dis}$ and the final loss $\mathcal{L}$ are formulated as follows:
\begin{equation} \label{eq:infonce}
    \mathcal{L}_{dis} = -\log \frac{e^{(d(h,r,t)-\gamma)/\tau}} {e^{(d(h,r,t)-\gamma)/\tau} + \mathcal{S}'_{neg}},
\end{equation}
\begin{equation}
\mathcal{S}'_{neg}=\sum_{(h,r,t') \in IB}e^{d(h,r,t')/\tau}
\end{equation}
\begin{equation}
	\mathcal{L}= \mathcal{L}_{cl}+ \beta \cdot \mathcal{L}_{dis}.
\end{equation}
Where $\beta$ represents the weights, they serve as the basis for ranking in model inference. We conduct a comprehensive empirical study of the potential ranking score options based on $\beta$ in the
supplementary materials.

\section{Experiment}

\subsection{Experimental Setup}
\subsubsection{Datasets}
We conducted evaluations on three different datasets to assess the effectiveness of our approach: WN18RR~\cite{Dettmers2017Convolutional2K}, FB15k-237~\cite{Ma2015KnowledgeGI} and OpenBG500~\cite{Deng2022ConstructionAA}. The statistics are shown in Table~\ref{tab:dataset}. The WN18RR and FB15k-237 datasets have been carefully constructed to eliminate inverse relations and prevent test set leakage. WN18RR is a dataset that includes $\sim 41k$ synsets and 11 relations from WordNet~\cite{miller1995wordnet}. FB15k-237 comprises $\sim 15k$ entities and 237 relations, extracted from Freebase. Lastly, we assessed our approach using OpenBG500, a unimodal Knowledge Graph (KG) comprising 249,743 entities derived from a deployed system, designed to serve as an open business knowledge repository. For text descriptions, we utilized the data provided by KG-BERT\cite{yao2019kg} for WN18RR and FB15k-237 datasets. For OpenBG500 datasets, we use GPT-3.5\cite{Chen2020BigSM} to generate the descriptions.

\begin{table}[ht]
\centering
\scalebox{0.77}{\begin{tabular}{@{}l|lllll@{}}
\toprule
Dataset          & \#Ent & \#Rel & \#Train & \#Dev & \#Test \\ \midrule
WN18RR           &   $40,943$  &    $11$    &  $86,835$   &   $3034$   &  $3134$     \\
FB15k-237        &   $14,541$  &    $237$  &  $272,115$  &  $17,535$ & $20,466$ \\
OpenBG500   &   $249,743$   &   $500$ &  $1,242,550$  & $5,000$  &  $5,000$  \\ 
\bottomrule
\end{tabular}}
\caption{Overall statistics of benchmark datasets.}
\label{tab:dataset}
\end{table}

\subsubsection{Baselines} 
We compare our methods with structure-based and description-based methods. The structure-based methods include TransE, DistMult, ComplEx, RotatE, TuckER, and CSPromp-KG. The description-based methods include KG-BERT, MTL-KGC, StAR, C-LMKE, KG-S2S, and SimKGC.

\subsubsection{Evaluation Metrics}
Based on previous research, we evaluate MoCoSA through the entity ranking task. For each test triplet $(h, r, t)$, the trained model aims at ranking all entities related to the given tail entity prediction pairs $(h,r)$ or head entity prediction pairs $(t,r^{-1})$ to predict $t$ or $h$. We employ four automated metrics: Mean Reciprocal Rank (MRR) and Hits@k (H@k for brevity), where $k$ takes on values from the set $\left\{1, 3, 10\right\}$. MRR is the average reciprocal rank of all test triplets. H@k calculates the proportion of correctly ranked entities within the top-k entities. All metrics are reported under the filtering settingas~\cite{Bordes2013TranslatingEF} that ignores the scores of all known true triplets in the training, evaluation, and test datasets and averaging in two directions: tail entity prediction and head entity prediction. In particular, we report results by submitting them to the official evaluation website\footnote{\url{https://tianchi.aliyun.com/OpenBG}} for OpenBG500.

\subsubsection{Hyperparameters}

The PLMs for the WN18RR and FB15k-237 are initialized with \emph{bert-base-uncased} and \emph{bert-base-chinese} for OpenBG500. It is expected that employing a superior PLM would be beneficial. The momentum queue size is 15360 and the number of hardest negatives is 192. The number of IRNS for each training instance is 3. Most hyperparameters are shared across all datasets, except for the learning rate and training epoch. We use grid search to determine the optimal learning rates within the range of $\left\{5e-4, 5e-5, 1e-5\right\}$. The initial temperature $\tau$ is set to 0.05, and the additive margin $\gamma$ is 0.02. The training process utilizes the AdamW optimizer with linear learning rate decay. The models are trained using a batch size of 768 on 1 A100 GPU. The training epochs for the WN18RR, FB15k-237, and OpenBG500 datasets are 100, 10, and 10, respectively.
\begin{table}[ht]
\centering
\scalebox{0.69}{\begin{tabular}{l|cccc}
\hline
\multirow{2}{*}{\bf Method} & \multicolumn{4}{c}{\bf OpenBG500}  \\ \cline{2-5}
                        & \bf MRR & \bf H@1 & \bf H@3 & \bf H@10 \\ \hline
\multicolumn{3}{l}{\textit{structure-based methods}}         \\ \hline
TransE ~\citep{bordes2013translating}$^\dagger$ & 30.4 & 20.7 &34.0 &51.3   \\
TransH ~\citep{Wang2014KnowledgeGE}$^\dagger$ & 29.6 & 14.3 &40.2 &56.9   \\
DistMult ~\citep{yang2014embedding}$^\dagger$ & 12.9 & 6.8 &13.1 &25.5   \\
ComplEx ~\citep{trouillon2016complex}$^\dagger$  & 15.6 &8.1 &18.7 & 31.3 \\ 
TuckER~\citep{Balazevic2019TuckERTF}$^\dagger$ & 54.1 & 42.8 & 61.5 & 73.5\\ \hline
\multicolumn{3}{l}{\textit{description-based methods}}         \\ \hline
KG-BERT ~\citep{yao2019kg}$^\dagger$ & 13.8 &7.1 &14.5 & 26.2 \\
GenKGC ~\citep{Xie2022FromDT}$^\dagger$ & - & 20.3 & 28.0 & 35.1 \\
C-LMKE ~\citep{Wang2022LanguageMA} &39.3 &29.6 &46.6 &58.4\\
SimKGC ~\citep{Wang2022SimKGCSC} &42.4 &30.7 &50.2 &67.9\\ \hline
\textbf{MoCoSA (Ours)}  & \textbf{63.4} & \textbf{53.1} & \textbf{71.1} & \textbf{83.0}\\ \hline
\end{tabular}}
\vspace{-5pt}
\caption{Results on the OpenBG500 dataset.
$^\dagger$: results are from ~\citet{Deng2022ConstructionAA}.}
\label{tab:openbg}
\vspace{-10pt}
\end{table}
\begin{table*}[ht]
\centering
\scalebox{0.9}{\begin{tabular}{l|cccc|cccc}
\hline
\multirow{2}{*}{\bf Method} & \multicolumn{4}{c|}{\bf WN18RR}      & \multicolumn{4}{c}{\bf FB15k-237}  \\ \cline{2-9}
                        & \bf MRR & \bf H@1 & \bf H@3 & \bf H@10 & \bf MRR & \bf H@1 & \bf H@3 & \bf H@10 \\ \hline
\multicolumn{9}{l}{\textit{structure-based methods}}         \\ \hline
TransE ~\citep{bordes2013translating}$^\dagger$ &  24.3 & 4.3  & 44.1 & 53.2 & 27.9 & 19.8 & 37.6 & 44.1 \\
DistMult ~\citep{yang2014embedding}$^\dagger$   &  44.4 & 41.2 & 47.0 & 50.4  &  28.1 & 19.9 & 30.1 & 44.6 \\
ComplEx ~\citep{trouillon2016complex}$^\dagger$ &  44.9 & 40.9  & 46.9 & 53.0 & 27.8 & 19.4 & 29.7 & 45.0 \\
RotatE ~\citep{sun2018rotate}$^\dagger$  & 47.6 & 42.8 & 49.2 & 57.1 & 33.8 & 24.1 &  37.5 & 53.3 \\
TuckER ~\citep{Balazevic2019TuckERTF}$^\dagger$ & 47.0 & 44.3 & 48.2 & 52.6 & 35.8 & 26.6 & 39.4 & 54.4 \\
CSPromp-KG  ~\citep{Chen2023DippingPS} & 57.5 & 52.2 & 59.6 & 67.8 & 35.8 & 26.9 & 39.3  & 53.8 \\ \hline
\multicolumn{9}{l}{\textit{description-based methods}}         \\ \hline
KG-BERT ~\citep{yao2019kg} & 21.6 & 4.1 &  30.2 & 52.4 &  -  &  - &  - &  42.0 \\
MTL-KGC ~\citep{kim-etal-2020-multi} & 33.1 & 20.3 & 38.3 & 59.7 & 26.7 & 17.2 & 29.8 & 45.8 \\
StAR ~\citep{wang2021structure} &  40.1 & 24.3 &  49.1 & 70.9 &  29.6 & 20.5 & 32.2 & 48.2 \\ 
C-LMKE ~\citep{Wang2022LanguageMA} & 59.8 & 48.0 & 67.5 & 80.6 & \textbf{40.4} & \textbf{32.4} & \textbf{43.9} & 55.6 \\
KG-S2S ~\citep{Chen2022KnowledgeIF} & 57.4 & 53.1 & 59.5 & 66.1 & 33.6 & 25.7 & 37.3 & 49.8 \\
SimKGC ~\citep{Wang2022SimKGCSC} & 67.1 & 58.7 & 73.1 & 81.7 & 33.3 & 24.6 & 36.2  & 51.0 \\ \hline
\textbf{MoCoSA (Ours)}  & \textbf{69.6} & \textbf{62.4} & \textbf{73.7} & \textbf{82.0} & 38.7 & 29.2 & 42.0  & \textbf{57.8} \\ \hline
\end{tabular}}
\caption{Main results on WN18RR and FB15k-237 datasets.
$^\dagger$: results are from~\citet{wang2021structure}.}
\label{tab:wn_and_fb}
\vspace{-10pt}
\end{table*}

\subsection{Main Result}
We conducted the analysis of our methodologies with previously established approaches to the link prediction task. Experimental results in Table~\ref{tab:openbg} and Table~\ref{tab:wn_and_fb} show that compared with the description-based methods and embedding-based methods. MoCoSA achieves state-of-the-art performance on WN18RR and OpenBG500. It achieves MRR 69.6\% on WN18RR in Table~\ref{tab:wn_and_fb} and 63.4\% on OpenBG500 in Table~\ref{tab:openbg}, beating all other methods achieving SOTA. This demonstrates the importance of incorporating structural information into PLMs and using effective negative sampling for KGC. Although MoCoSA slightly lags behind on the FB15k237 dataset, (MRR 38.7\% vs 40.4\%), it is more competitive compared to other methods and improves the H@10 from 55.6\% to 57.8\%.  

We investigated the effectiveness of the adaptable structural encoder and the two negative sampling strategies. As shown in Table~\ref{tab:ablation study}, our experiments demonstrated that MoCoSA$_\textbf{IB+MH}$ resulted in significant performance improvements in MRR and Hits@$\left\{1, 3, 10\right\}$ both on three datasets. MoCoSA$_\textbf{IB+ASE}$ makes MRR imrpove from 68.8\% to 69.0\% on WN18RR but makes MRR improve from 37.1\% to 38.7\% on FB15k-237. This discrepancy can be explained by Cartesian Product Relations(CPR)~\cite{Chen2022KnowledgeIF, Lv2022DoPM, Akrami2020RealisticRO}. On WN18RR, the structural relationships are derived from the entities' definitions, establishing a direct mapping between structural information and textual descriptions. Consequently, the descriptions support the inference of structural information, resulting in less improvement when adding ASE. However, entities represent in the real world and the descriptions only partially facilitate inference on FB15k-237. Excessive reliance on textual information may have detrimental effects on performance. Conversely, incorporating structural information can alleviate this reliance.

Furthermore, MoCoSA$_\textbf{IB+IR}$ incorporates intra-relation negative sampling during the training process and generally improves the H@$\left\{3, 10\right\}$ metric with the inclusion of three instra-relation negative samples on WN18RR and OpenBG500. This suggests that it is more effective in capturing semantic distinctions between different entities under the same relation during training. Each strategy exhibits varying efficacy across different datasets. Despite the potential drawbacks associated with the different negative sampling strategies, the aggregated negative sampling strategy yields superior performance improvement.
\begin{table}[!htbp]
	\centering
        \resizebox{0.48\textwidth}{!}{
	   \begin{tabular}{lccccc}
		\toprule
                \bf Method & \bf PLM & \bf T/Ep & \bf Inf \\
		\midrule\midrule
            KG-BERT~\cite{yao2019kg} & $RoBERTa_{base}$ & 79m & 954m  \\
            StAR~\cite{Wang2020StructureAugmentedTR} & $RoBERTa_{base}$ &42m &27m  \\
            GenKGC~\cite{Xie2022FromDT} & $BART_{base}$ &5m & 88m \\
            KG-S2S~\cite{Chen2022KnowledgeIF} & $T5_{base}$ &10m &81m  \\
            SimKGC~\cite{Wang2022SimKGCSC} & $BERT_{base}$ & 6m & 2m  \\
                        \midrule
            \textbf{MoCoSA (Ours)} & $BERT_{base}$ & 7m & 2m \\
            \bottomrule
            \vspace{-10pt}
            \end{tabular}
        }
        \caption{
        Efficiency comparisons between MoCoSA and other PLM-based methods on WN18RR. Here, T/Ep and Inf indicate the training time per epoch and the inference time.
        }
        \vspace{-10pt}
        \label{tab:time} 
\end{table}
We calculate the specific time for training and inference incurred across on WN18RR. As shown in Table~\ref{tab:time}, MoCoSA requires less time for a single epoch compared to most description-based methods, while also using faster inference times. Here, we just want to reiterate the significance of inference efficiency when designing novel models.

\subsection{Intrinsic Analysis}

\begin{figure}[t]
    \centering
    \begin{subfigure}{0.234\textwidth}
        \centering
        \includegraphics[width=1.0\linewidth]
        {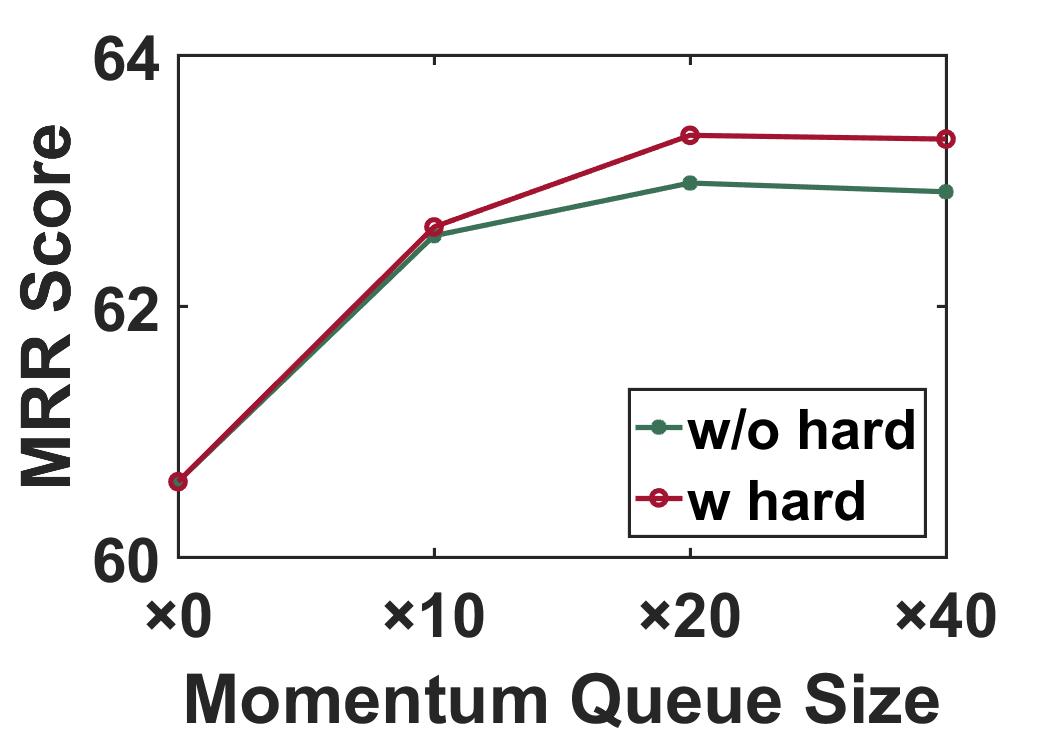}
        \caption{}
        \label{fig:curve_MH}
    \end{subfigure}
    \begin{subfigure}{0.233\textwidth}
        \centering
        \includegraphics[width=1.0\textwidth]
        {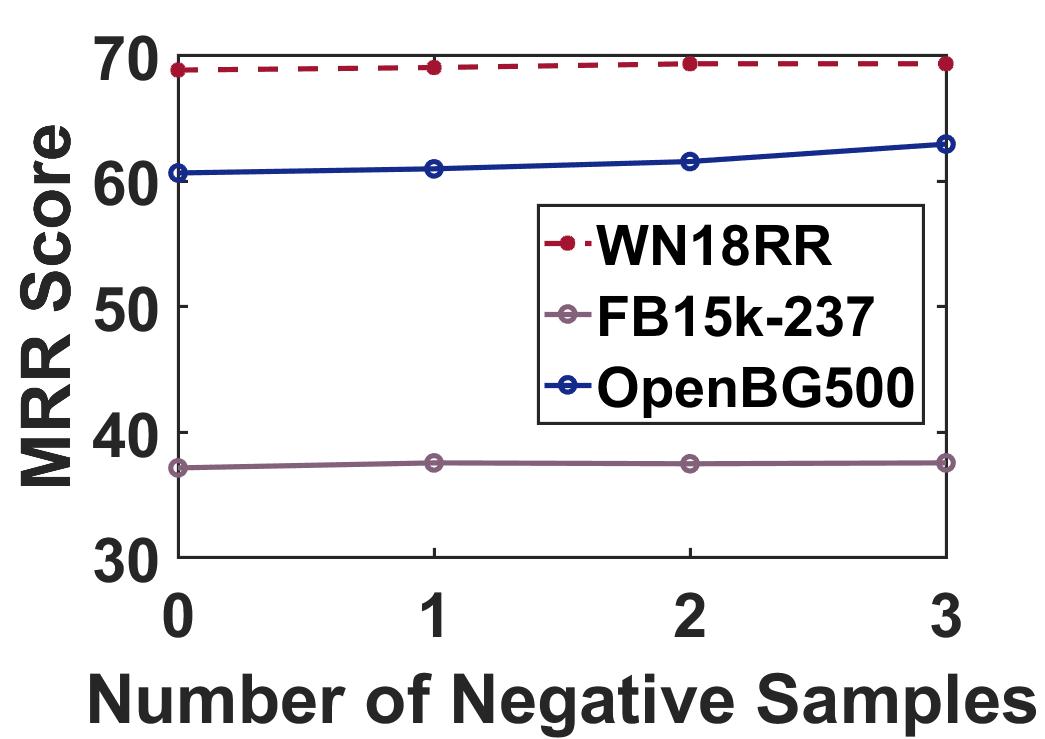}
        \caption{}
        \label{fig:curve_IR}
    \end{subfigure}
    \caption{(a) the effect of the size of momentum queue size on OpenBG500. w/o hard means not using a mix-up strategy. Ticks on the horizontal axis represent multiples of the batch size. (b) the effect of the number of IR negatives.} 
    \label{fig: curve}
\end{figure}

\subsubsection{The Analysis of the Momentum Hard Negative Sampling}
In order to evaluate the efficacy of momentum hard (MH) negative sampling, we conducted experiments on OpenBG500 by adjusting the size of the momentum queue to control the number of negative samples and investigating the impact of employing a mixing approach on the negative samples. The results are depicted in Figure~\ref{fig:curve_MH}, which demonstrates the efficacy of enlarging the momentum queue to accommodate an increased number of negative samples. Considering computational constraints, we made a judicious choice regarding the queue size. Notably, the utilization of a mixing strategy for the most challenging negative samples within the momentum queue was found to significantly enhance the quality of the negative samples and improve the model performance.

\subsubsection{The Effect of the Number of Intra-relation Negatives}
To assess the effectiveness of intra-relation (IR) negative sampling, we conducted experiments by varying the number of intra-relational negatives from 0 to 3 on WN18RR, FB15K-237, and OpenBG500. Figure ~\ref{fig:curve_IR} illustrates that increasing the number of negative samples enhances the model's performance significantly on WN18RR and FB15k-237. However, our research has observed a limited number of these strategies, primarily due to computational overhead constraints. Consequently, enhancing the accessibility of IR negatives will be a primary focus of future research.

\begin{table}[!htbp]
	\centering
        \resizebox{0.4\textwidth}{!}{
	   \begin{tabular}{lccccc}
		\toprule
                \bf $ASE(\cdot)$ & \bf PLM & \bf MRR & \bf H@1 & \bf H@3 & \bf H@10 \\
		\midrule\midrule
             \multirow{2}{*}{$\textbf{h}+\textbf{t}$} & $\quad \times$ &8.2 & 3.4 & 9.7 & 17.2 \\
             & $\quad \checkmark$ &38.0 &28.8 &41.3 &56.5  \\
             \midrule
            \multirow{2}{*}{$\textbf{h} \odot \textbf{t}$} & $\quad \times$ &15.2 &10.7 &16.5 &23.8  \\
             & $\quad \checkmark$ &38.7 & 29.2 &42.0 &57.8 \\ 
            \bottomrule
            \end{tabular}
        }
        \caption{
        Results on FB15k-237 with three map functions for the structural encoder. $\checkmark$ means unifying PLM for training, while $\times$ means only using the structural encoder for training. $\odot$ denotes the Hadamard product.
        }
        \vspace{-10pt}
        \label{tab:model flexibility} 
\end{table}

\subsubsection{Flexibility in Adaptable Structural Encoder}

In order to evaluate the adaptation of the structural-based methods, we incorporated two structure-based methods as the mapping function in $ASE(\cdot)$. As shown in Table~\ref{tab:model flexibility}, We observed that as the performance of the structural method improves, the fusion with the PLM led to even greater performance gains. This reinforces the notion that a well-performing structure-based method enhances the overall capabilities of PLMs in the KGC task. Therefore, it is also important to leverage high-quality structural information for improved results in the joint process. Additionally, we demonstrate more structural-based methods in the supplementary materials.

\subsubsection{The Effect of Relation-based Re-ranking}

\begin{figure}[t]
    \centering
    \begin{subfigure}{0.234\textwidth}
        \centering
        \includegraphics[width=1.0\linewidth]
        {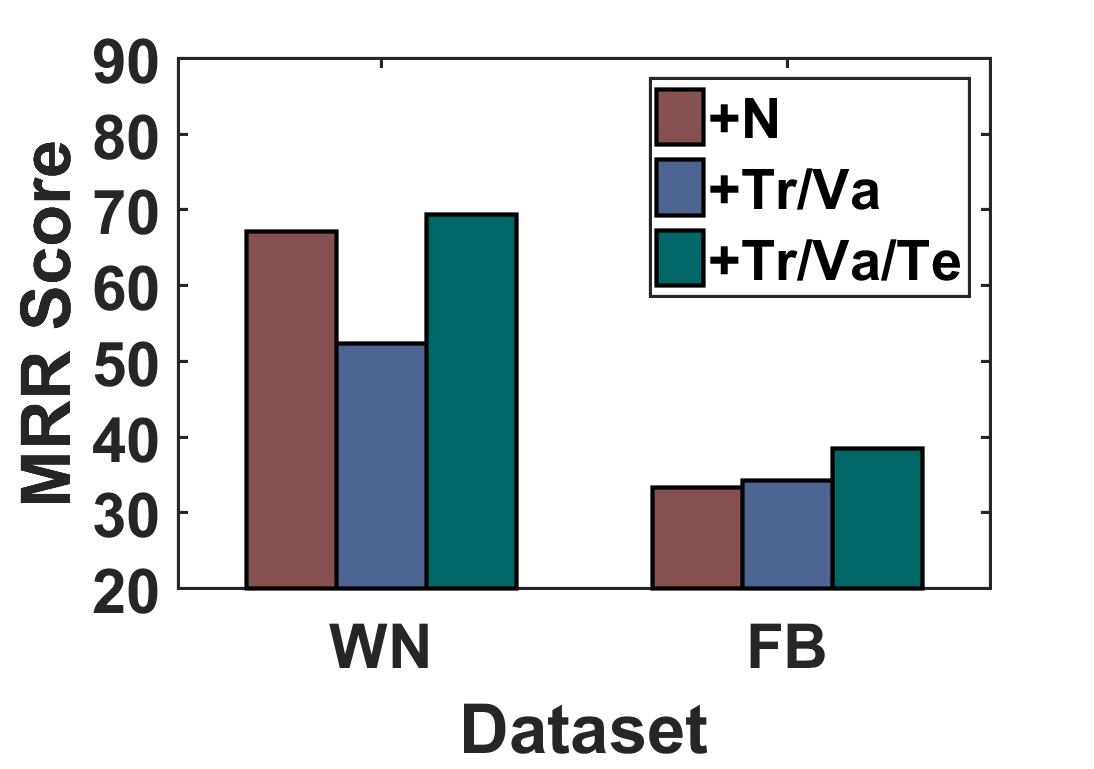}
        \caption{}
        \label{fig:bar_MRR}
    \end{subfigure}
    \begin{subfigure}{0.229\textwidth}
        \centering
        \includegraphics[width=1.0\textwidth]
        {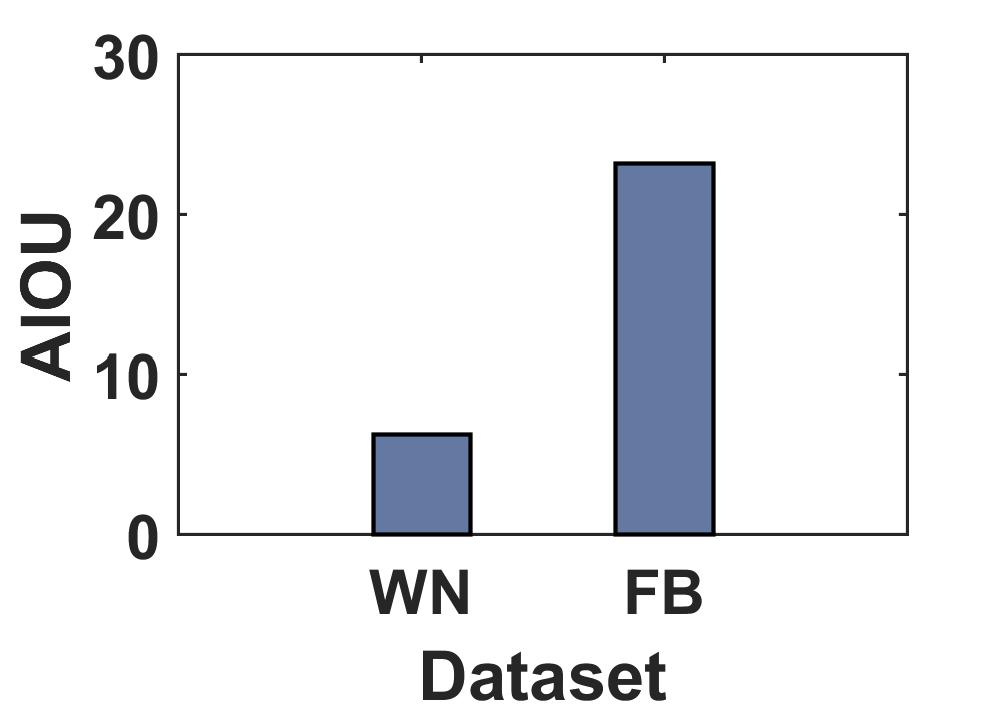}
        \caption{}
        \label{fig:bar_AIOU}
    \end{subfigure}
    \caption{(a) the effect of relation-based re-ranking in WN(WN18RR), FB(FB15k-237). N means not incorporating any existing knowledge from the dataset. Tr, va, and Te mean using knowledge in training, validation, and test set, respectively. (B) statistics of AIOU on WN18RR and FB15k-237}
    \label{fig:bar_curve}
    \vspace{-5pt}
\end{figure}

To validate the effectiveness of the relation-based re-ranking strategy, we conducted experiments as shown in Figure~\ref{fig:bar_MRR}. This strategy demonstrates performance improvements when leveraging existing knowledge in the training, validation, and test sets both on WN18RR and FB15k-237. On WN18RR, employing existing knowledge solely in the training set leads to a drop in MRR. We conducted an in-depth analysis to understand the reasons behind this observation, as shown in Figure~\ref{fig:bar_AIOU}. Compared to FB15k-237, the WN18RR dataset shows a fact between triples that leads to lower average intersection over union (AIOU) values between the training and test sets for tail entities. Consequently, limiting the re-ranking scope solely to the training set has a detrimental effect.

\begin{table*}[ht]
\centering
\scalebox{0.8}{\begin{tabular}{l|cccc|cccc|cccc}
\hline
\multirow{2}{*}{\bf Method} & \multicolumn{4}{c|}{\bf WN18RR}      & \multicolumn{4}{c|}{\bf FB15k-237}   &
\multicolumn{4}{c}{\bf OpenBG500} \\ \cline{2-13}
                        & \bf MRR & \bf H@1 & \bf H@3 & \bf H@10 & \bf MRR & \bf H@1 & \bf H@3 & \bf H@10 & \bf MRR & \bf H@1 & \bf H@3 & \bf H@10\\ \hline
MoCoSA$_\textbf{IB}$  & 68.8 & 61.6 & 72.8 & 81.2 & 37.1 & 28.0 & 40.2  & 55.6 & 60.6 &49.2 &69.7 &82.1\\
MoCoSA$_\textbf{IB+MH}$  & \textbf{69.6} & \textbf{62.4} & 73.7 & 82.0 & 37.4 & 28.4 & 40.5  & 55.9  & \textbf{63.4} & \textbf{53.1} &71.1 &83.0\\
MoCoSA$_\textbf{IB+ASE}$  & 69.0 & 61.7 & 73.1 & 81.4 & \textbf{38.7} & \textbf{29.2} & \textbf{42.0}  & \textbf{57.8} & 62.8 &52.0 &71.1 &83.2\\
MoCoSA$_\textbf{IB+IR}$  & 69.3 & 61.2 & \textbf{74.8} & \textbf{83.5} & 37.5 & 28.4 & 40.8  & 55.8 & 62.9 &51.9 &\textbf{71.4} &\textbf{83.4}\\
MoCoSA$_\textbf{IB+MH+ASE+IR}$  & 69.3 & 61.8 & 74.4 & 82.9 & 38.0 & 28.8 & 41.0  & 56.2 & 63.2 &52.5 &71.2 &82.8\\ \hline
\end{tabular}}
\vspace{-5pt}
\caption{Ablation study on WN18RR, FB15k-237 and OpenBG500. "IB", "MH", "ASE" and "IR" refer to in-batch negatives, momentum hard negatives, adaptable structural encoder, and instra-relation negatives.}
\label{tab:ablation study}
\vspace{-10pt}
\end{table*}

\section{Conclusion}

In this paper, we have presented MoCoSA, a novel structure-augmented pre-trained language model designed for link prediction tasks in knowledge graph completion. To better incorporate structural information, we propose the adaptable structure encoder for enabling the PLMs to acquire structural features. To optimize model training, we propose the incorporation of three distinct contrastive learning strategies: in-batch contrast, momentum contrast, and instra-relation contrast. Based on these strategies, MoCoSA outperforms state-of-the-art methods on WN18RR, FB15k-237, and OpenBG500 by conducting extensive evaluations.
 

\clearpage

\bibliography{main}

\end{document}